\documentclass{article} % For LaTeX2e
\usepackage{iclr2020_conference,times}

\usepackage{amsmath,amsfonts,bm}

\def\eqref#1{equation~\ref{#1}}
\def\floor#1{\lfloor #1 \rfloor}
\def\1{\bm{1}}

\def\rx{{\textnormal{x}}}

\def\vs{{\bm{s}}}

\def\vu{{\bm{u}}}
\def\vv{{\bm{v}}}

\def\vz{{\bm{z}}}

\def\evo{{o}}

\def\mA{{\bm{A}}}

\def\mK{{\bm{K}}}

\def\mO{{\bm{O}}}

\def\mQ{{\bm{Q}}}

\def\mV{{\bm{V}}}
\def\mW{{\bm{W}}}

\DeclareMathAlphabet{\mathsfit}{\encodingdefault}{\sfdefault}{m}{sl}
\SetMathAlphabet{\mathsfit}{bold}{\encodingdefault}{\sfdefault}{bx}{n}
\newcommand{\tens}[1]{\bm{\mathsfit{#1}}}

\def\tU{{\tens{U}}}

\def\tZ{{\tens{Z}}}

\def\emA{{A}}

\def\emO{{O}}

\def\emV{{V}}

\newcommand{\etens}[1]{\mathsfit{#1}}

\def\etO{{\etens{O}}}

\def\etU{{\etens{U}}}

\newcommand{\R}{\mathbb{R}}

\usepackage{hyperref}
\usepackage{url}
\usepackage{graphicx} % DO NOT CHANGE THIS
\usepackage{wrapfig} % DO NOT CHANGE THIS
\usepackage{lipsum} % DO NOT CHANGE THIS
\usepackage{adjustbox} % DO NOT CHANGE THIS
\usepackage{sidecap} % DO NOT CHANGE THIS
\usepackage{array} % DO NOT CHANGE THIS

\title{SesameBERT: Attention for Anywhere}

\author{\bf{Ta-Chun Su, Hsiang-Chih Cheng}\\
Cathay Financial Holdings Lab\\
\texttt{\{bgg, alberthcc\}@cathayholdings.com.tw} 
}

\begin{document}

\maketitle

\begin{abstract}
Fine-tuning with pre-trained models has achieved exceptional results for many language tasks. In this study, we focused on one such self-attention network model, namely BERT, which has performed well in terms of stacking layers across diverse language-understanding benchmarks. However, in many downstream tasks, information between layers is ignored by BERT for fine-tuning. In addition, although self-attention networks are well-known for their ability to capture global dependencies, room for improvement remains in terms of emphasizing the importance of local contexts. In light of these advantages and disadvantages, this paper proposes SesameBERT, a generalized fine-tuning method that (1) enables the extraction of global information among all layers through Squeeze and Excitation and (2) enriches local information by capturing neighboring contexts via Gaussian blurring. Furthermore, we demonstrated the effectiveness of our approach in the HANS dataset, which is used to determine whether models have adopted shallow heuristics instead of learning underlying generalizations. The experiments revealed that SesameBERT outperformed BERT with respect to GLUE benchmark and the HANS evaluation set.
\end{abstract}

\section{Introduction}

In recent years, unsupervised pretrained models have dominated the field of natural language processing (NLP). The construction of a framework for such a model involves two steps: pretraining and fine-tuning. During pretraining, an encoder neural network model is trained using large-scale unlabeled data to learn word embeddings; parameters are then fine-tuned with labeled data related to downstream tasks.\par
Traditionally, word embeddings are vector representations learned from large quantities of unstructured textual data such as those from Wikipedia corpora \citep{Mikolov2013}. Each word is represented by an independent vector, even though many words are morphologically similar. To solve this problem, techniques for contextualized word representation \citep{Peters2018,Devlin2019} have been developed; some have proven to be more effective than conventional word-embedding techniques, which extract only local semantic information of individual words. By contrast, pretrained contextual representations learn sentence-level information from sentence encoders and can generate multiple word embeddings for a word. Pretraining methods related to contextualized word representation, such as BERT \citep{Devlin2019}, OpenAI GPT \citep{Radford2018}, and ELMo \citep{Peters2018}, have attracted considerable attention in the field of NLP and have achieved high accuracy in GLUE tasks such as single-sentence, similarity and paraphrasing, and inference tasks \citep{Wang2019}. Among the aforementioned pretraining methods, BERT, a state-of-the-art network, is the leading method that applies the architecture of the Transformer encoder, which outperforms other models with respect to the GLUE benchmark. BERT's performance suggests that self-attention is highly effective in extracting the latent meanings of sentence embeddings.\par
This study aimed to improve contextualized word embeddings, which constitute the output of encoder layers to be fed into a classifier. We used the original method of the pretraining stage in the BERT model. During the fine-tuning process, we introduced a new architecture known as Squeeze and Excitation alongside Gaussian blurring with symmetrically SAME padding ("SESAME" hereafter). First, although the developer of the BERT model initially presented several options for its use, whether the selective layer approaches involved information contained in all layers was unclear. In a previous study, by investigating relationships between layers, we observed that the Squeeze and Excitation method \citep{Hu2017} is key for focusing on information between layer weights. This method enables the network to perform feature recalibration and improves the quality of representations by selectively emphasizing informative features and suppressing redundant ones.
Second, the self-attention mechanism enables a word to analyze other words in an input sequence; this process can lead to more accurate encoding. The main benefit of the self-attention mechanism method is its high ability to capture global dependencies. Therefore, this paper proposes the strategy, namely Gaussian blurring, to focus on local contexts. We created a Gaussian matrix and performed convolution alongside a fixed window size for sentence embedding. Convolution helps a word to focus on not only its own importance but also its relationships with neighboring words. Through such focus, each word in a sentence can simultaneously maintain global and local dependencies.\par

We conducted experiments with our proposed method to determine whether the trained model could outperform the BERT model. We observed that SesameBERT yielded marked improvement across most GLUE tasks. In addition, we adopted a new evaluation set called HANS \citep{McCoy2019}, which was designed to diagnose the use of fallible structural heuristics, namely the lexical overlap heuristic, subsequent heuristic, and constituent heuristic. Models that apply these heuristics are guaranteed to fail in the HANS dataset. For example, although BERT scores highly in the given test set, it performs poorly in the HANS dataset; BERT may label an example correctly not based on reasoning regarding the meanings of sentences but rather by assuming that the premise entails any hypothesis whose words all appear in the premise \citep{Dasgupta2018}. By contrast, SesameBERT performs well in the HANS dataset; this implies that this model does not merely rely on heuristics. In summary, our final model proved to be competitive on multiple downstream tasks.

\section{RELATED WORK}

\subsection{Unsupervised Pretraining in NLP}
Most related studies have used pretrained word vectors \citep{Mikolov2013,Pennington2014} as the primary components of NLP architectures. This is problematic because word vectors capture semantics only from a word's surrounding text. Therefore, a vector has the same embedding for the same word in different contexts, even though the word's meaning may be different.\par
Pretrained contextualized word representations overcome the shortcomings of word vectors by capturing the meanings of words with respect to context. ELMo \citep{Peters2018} can extract context-sensitive representations from a language model by using hidden states in stacked LSTMs. Generative pretraining \citep{Radford2018} uses the "Transformer encoder" rather than LSTMs to acquire textual representations for NLP downstream tasks; however, one limitation of this model is that it is trained to predict future left-to-right contexts of a unidirectional nature. BERT \citep{Devlin2019} involves a masked language modeling task and achieves high performance on multiple natural language-understanding tasks. In BERT architecture, however, because the output data of different layers encode a wide variety of information, the most appropriate pooling strategy depends on the case. Therefore, layer selection is a challenge in learning how to apply the aforementioned models. 

\subsection{Squeeze and Excitation}
The Squeeze and Excitation method was introduced by \citet{Hu2017}, who aimed to enhance the quality of representations produced by a network. Convolutional neural networks traditionally use convolutional filters to extract informative features from images. Such extraction is achieved by fusing the spatial and channel-wise information of the image in question. However, the channels of such networks' convolutional features have no interdependencies with one another. The network weighs each of its channels equally during the creation of output feature maps. Through Squeeze and Excitation, a network can take advantage of feature recalibration and use global information to emphasize informative features and suppress less important ones.

\subsection{Localness Modeling}
The self-attention network relies on an attention mechanism to capture global dependencies without considering their distances by calculating all the positions in an input sequence. Our Gaussian-blurring method focuses on learning local contexts while maintaining a high ability to capture long-range dependencies. Localness modeling was considered a learnable form of Gaussian bias \citep{Yang2019} in which a central position and dynamic window are predicted alongside intermediate representations in a neural network. However, instead of using Gaussian bias to mask the logit similarity of a word, we performed Gaussian bias in the layer after the embedding layer to demonstrate that performing element-wise operations in this layer can improve the model performance.

\subsection{Diagnosing Syntactic Heuristics}
A recent study \citep{McCoy2019} investigated whether neural network architectures are prone to adopting shallow heuristics to achieve success in training examples rather than learning the underlying generalizations that need to be captured. For example, in computer vision, neural networks trained to recognize objects are misled by contextual heuristics in cases of monkey recognition \citep{Wang2017}. For example, in the field of natural language inference (NLI), a model may predict a label that contradicts the input because the word "not", which often appears in examples of contradiction in standard NLI training sets, is present \citep{Naik2018,Carmona2018}. In the present study, we aimed to make SesameBERT robust with respect to all training sets. Consequently, our experiments used HANS datasets to diagnose some fallible structural heuristics presented in this paper \citep{McCoy2019}.

%% Please note that we have introduced automatic line number generation
%% into the style file for \LaTeXe. This is to help reviewers
%% refer to specific lines of the paper when they make their comments. Please do
%% NOT refer to these line numbers in your paper as they will be removed from the
%% style file for the final version of accepted papers.

\section{Methods}
We focused on BERT, which is the encoder architecture of a multilayer Transformer \citep{Vaswani2017}, featuring some improvements. The encoder consists of L encoder layers, each containing two sublayers, namely a multihead self-attention layer and a feed-forward network. The multihead mechanism runs through a scaled dot product attention function, which can be formulated by querying a dictionary entry with key value pairs \citep{Miller2016}. The self-attention input consists of a query $\mQ \in { \R^{l\times d}}$, a key $\mK\in {\R^{l\times d}}$, and a value $\mV \in {\R^{l\times d}}$, where $l$ is the length of the input sentence, and $d$ is the dimension of embedding for query, key and value. For subsequent layers, $\mQ$, $\mK$, $\mV$ comes from the output of the previous layer. The scaled dot product attention \citep{Vaswani2017} is defined as follows:
\begin{equation}
    Attention(\mQ,\mK,\mV) = softmax(\frac{\mQ \mK^{T}}{\sqrt{d}})\cdot \mV
\end{equation}
The output represents the multiplication of the attention weights $\mA$ and the vector $\vv$, where $\mA =softmax(\frac{\mQ \mK^{T}}{\sqrt{d}}) \in \R^{l\times l}$. The attention weights $\emA_{i,j}$ enabled us to better understand about the importance of the $i$-th key-value pairs with respect to the $j$-th query in generating the output \citep{Bahdanau2015}. During fine-tuning, We used the output encoder layer from the pretrained BERT model to create contextualized word embeddings and feed these embeddings into the model. Although several methods have been developed for extracting contextualized embeddings from various layers, we believed that these methods had substantial room for improvement. Therefore, we used Squeeze and Excitation to solve the aforementioned problem.

\begin{figure*}[t]
        \begin{center}
        \includegraphics[scale=0.23]{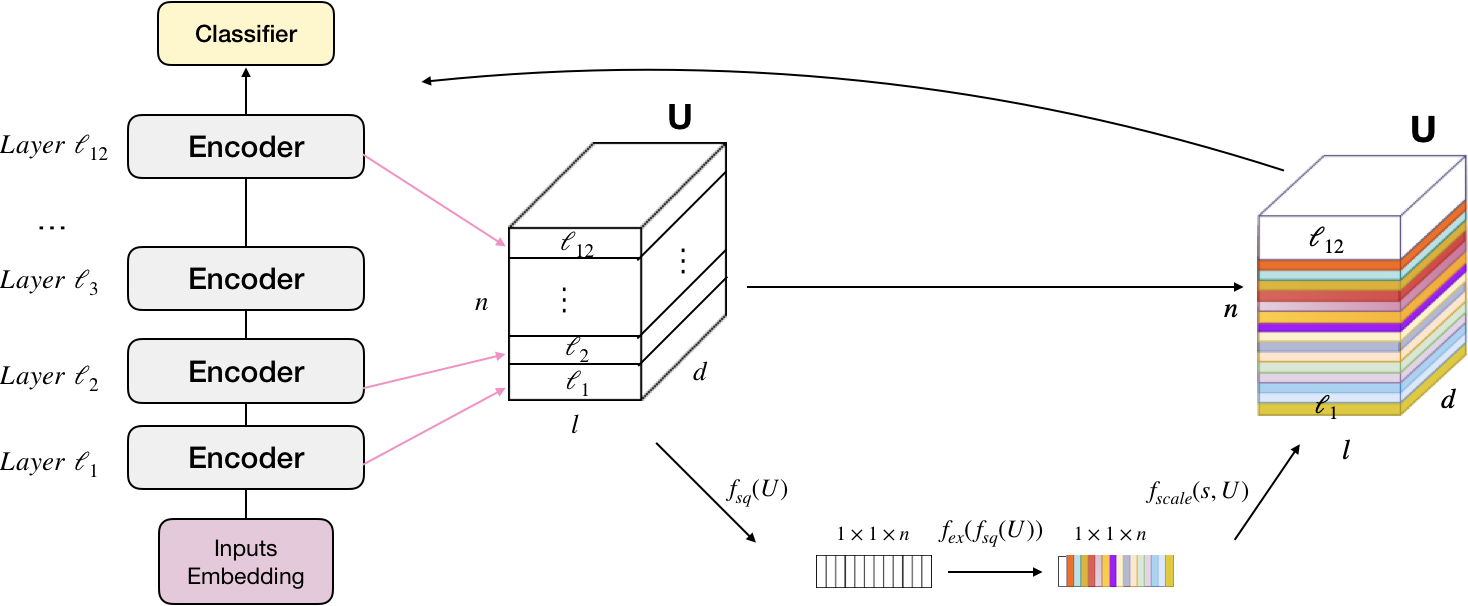}
        \caption{\label{Figure1}We extracted the output from each layer of the encoders and concatenated all the layers to form a three-dimensional tensor $\tU$. We then performed Squeeze $f_{sq}(\tU)$ and Excitation $f_{ex}(f_{sq}(\tU))$ to obtain the weight of each output layer. Finally, we fed the weighted average of all layers into the classifier. In this work we employed $n$ = 12 attention layers.}
        \end{center}
\end{figure*} 

\subsection{Squeeze and Excitation}
In this study, we proposed the application of Squeeze and Excitation \citep{Hu2017}; its application to the output of the encoder layer was straightforward once we realized that the number of channels was equivalent to the number of layers. Therefore, we intended to use the term channels and layers interchangeably. \par

First, we defined $\tU_{:,:,k}$ as the output of the $k$-th encoder layer, for all $1 \leq k \leq n$. We wanted to acquire global information from between the layers before feeding the input into the classifier; therefore, we concatenated all the output from each encoder layer to form the feature maps $\tU \in \R^{l\times d\times n}$. In the squeeze step, by using global average pooling on the $k$th layer, we were able to squeeze the global spatial information into a layer descriptor. In other words, we set the $k$th layer's output of the squeeze function as $\tZ_{:,:,k}$.
\begin{equation}
\tZ_{:,:,k} = f_{sq}(\tU_{k}) = \frac{1}{l\times d}\sum_{i=1}^l \sum_{j=1}^d \etU_{i,j,k}
\end{equation} 
In the excitation step, we aimed to fully capture layer-wise dependencies. This method uses the layer-wise output of the squeeze operation $f_{sq}$ to modulate interdependencies of all layers. Excitation is a gating mechanism with a sigmoid activation function that contains two fully connected layers. Let $\mW_{1}$ and $\mW_{2}$ be the weights of the first and second fully connected layers, respectively, and let $r$ be the bottleneck in the layer excitation that encodes the layer-wise dependencies; therefore, $\mW_{1} \in \R^{n\times \frac{n}{r}}$, and $\mW_{2} \in \R^{\frac{n}{r} \times n}$. The excitation function $f_{ex}$:

\begin{equation}
    \vs = f_{ex}(\vz)=\sigma(ReLU(\vz, \mW_{1}), \mW_{2})
\end{equation}
where $\vz$ is the vector squeezed from tensor $\tZ$. \par

Finally, we rescaled the output $\tZ_{:,:,k}$ by multiplying it by $\vs_k$. The rescaled output is deonted as $\widetilde{\vu}_{k}$. The scaling function $f_{scale}$ is defined as follows:
\begin{align}
    \widetilde{\vu}_{k} &= f_{scale}(\vs_{k}, \tU_{:,:,k})
\end{align}
We concatenated all rescaled outputs from all encoder layers to form our rescaled feature maps $\widetilde{\vu}$. The architecture is shown in Figure \ref{Figure1}. We then extracted layers from the rescaled feature maps, or calculated a weighted average layer $\widetilde{\vu}_{avg}$.
\begin{align}
    \widetilde{\vu}_{avg} = \frac{\sum_{k=1}^{n}f_{scale}(\vs_k, \tU_{:,:,k})}{\sum_{k=1}^{n}\vs_{k}}
\end{align}

\subsection{Gaussian Blurring}
Given an input sequence $X=\{x_{1}, x_{2}, ...,x_{l}\}\in \R^ {l\times d}$, the model transformed it into queries $\mQ$, keys $\mK$, and values $\mV$, where $\mQ, \mK$, and $\mV \in \R ^{l\times d}$. Multihead attention enabled the model to jointly attend to information from different representation subspaces at different positions. Thus, the three types of representations are split into h subspaces of size $\frac{d}{h}$ to attend to different information. For example, $\mQ=(\mQ^{1}, \mQ^{2},..., \mQ^{h})$ with $\mQ^{i}\in \R ^{l\times \frac{d}{h}}$ for all $1 \leq i \leq h$. In each subspace h, the element $\evo_{i}^{h}$ in the output sequence $\mO^{h}=(\evo_{1}^{h}, \evo_{2}^{h}, ...,\evo_{l}^{h})$ is computed as follows:
\begin{equation}
    \evo_{i}^{h} = Attention(q_{i}^{h}, \mK^{h})\mV^{h}
\end{equation}
where $\evo_{i}^{h}\in \R ^{\frac {d}{h}}$.\par
To capture the local dependency related to each word, we first used a predefined fixed window size $k$ to create a Gaussian blur $g$, where $g\in \R ^{k}$:
\begin{equation}
g(\rx; \sigma, k) = exp(\frac{-(\rx-\floor{\frac{k}{2}})^2}{2\sigma^2})
\end{equation}
where $\sigma$ refers to the standard deviation. Several Gaussian-blurring strategies are feasible for applying convolutional operations to attention outputs. 

\subsubsection{Gaussian Blurring on Attention Outputs}

\begin{SCfigure}
  \caption{\label{Figure2}Diagram of a one-dimensional Gaussian blur kernel, which was convoluted through the input dimension. This approach enabled the central word to acquire information concerning neighboring words with weights proportional to the Gaussian distribution.}
  \includegraphics[width=0.5\textwidth]{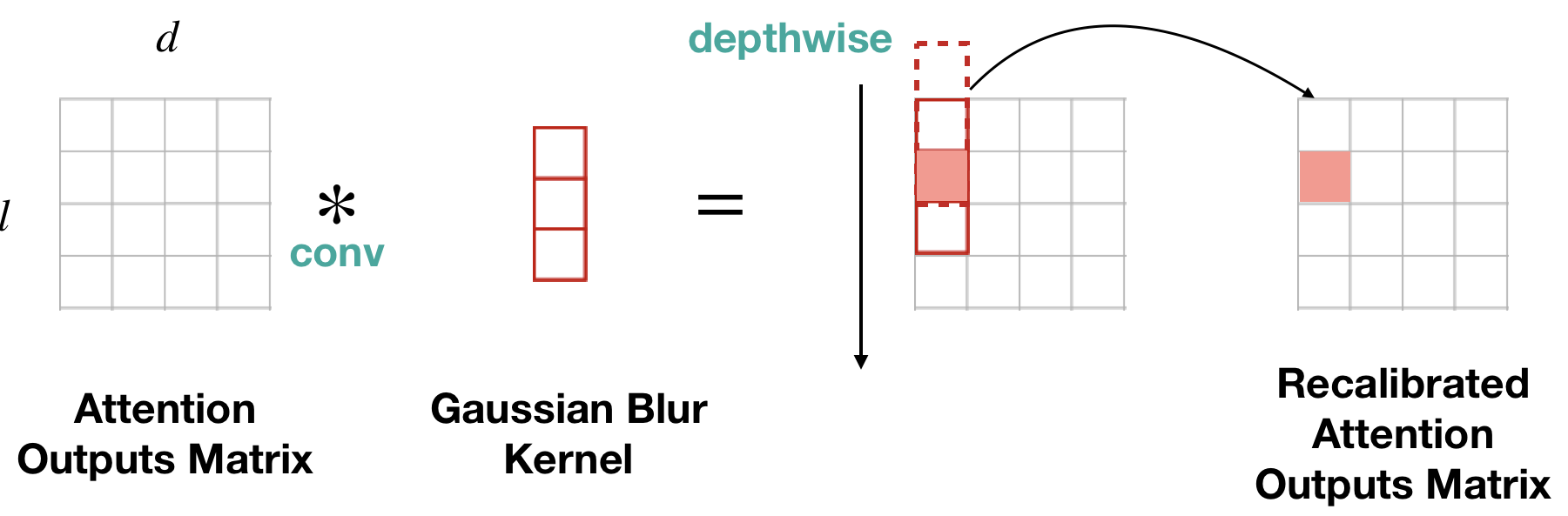}
\end{SCfigure}

The first strategy focuses on each attention output $\mO^{h}$. We restrict $\hat{\etO}_{i,j,:}^{h}$ to a local scope with a fixed size $k$ centered at the position $i$ and dimension $j$, where $1 \leq j \leq d$, and k can be any odd number between $1$ and $l$, expressed as follows:
\begin{equation}
\hat{\etO}_{i,j,:}^{h} = [\emO^{h}_{i-\floor{\frac{k}{2}},j},...,\emO^{h}_{i,j}  ...,\emO^{h}_{i+\floor{\frac{k}{2}},j}]
\end{equation}
We then enhance the localness of $\hat{\etO}_{i,j,:}^{h}$ through a parameter-free $1D$ convolution operation with $g$. 

\begin{equation}
\widetilde{\emO}_{i,j}^{h}= \hat{\etO}_{i,j,:}^{h} \cdot g
\end{equation}
The final attention output is $\widetilde{\mO}^{h}$, which is the dot product between the Gaussian kernel and the corresponding input array elements at every position of $\hat{\etO}_{i,j,:}^{h}$,

\begin{equation}
\widetilde{\mO}^{h}= \mO^{h}\ast g
\end{equation}
where $\ast$ is defined as a convolution operation, as illustrated in Figure \ref{Figure2}. 

More specifically, $\widetilde{\emO}^{h}_{i,j}$, the entry of $\widetilde{\emO}^{h}$ in the $i$-th row and $j$-th column, equals $blur(\emO^{h}_{i,j})$:
\begin{align}
    \widetilde{\emO}^{h}_{ij} & = blur(\emO^{h}_{i,j}) \nonumber\\
    & = \sum_{\rx \in [-k, k]} g(\rx;\sigma, k)\emO_{i+\rx,j} \nonumber\\
    & = \sum_{\rx \in [-k, k]} g(\rx; \sigma, k)\sum_l \emA_{i+\rx, l}\emV_{l,j}
\end{align}

\subsubsection{Gaussian Blurring on Values}
Another option focuses on values V. We applied the aforementioned method again but restrict $\mV^{h}$ to a local scope. The final attention output $\widetilde{O}^{h}$ is denoted as follows:
\begin{equation}
\widetilde{\mO}^{h}= Attention(\mQ^h, \mK^h) (\mV^{h} \ast g)
\end{equation}
The difference between the present method and the method of performing Gaussian blurring on attention outputs and values is that the method of performing Gaussian blurring on attention outputs and values places greater emphasis on the interaction of cross-query vectors, whereas the present method focuses on cross-value vectors.
Finally, the outputs of the h attention heads are concatenated to form the final output representation $\widetilde{\mO}$:
\begin{equation}
    \widetilde{\mO} = (\widetilde{\mO}^{1}, \widetilde{\mO}^{2}, ..., \widetilde{\mO}^{h})
\end{equation}
where $\widetilde{\mO}\in \R^{l\times d}$.
The multihead mechanism enables each head to capture distinct linguistic input properties \citep{Li2019}. Furthermore, because our model is based on BERT, which builds an encoder framework with a stack of 12 layers, we were able to apply locality modeling to all layers through Squeeze and Excitation. Therefore, we expected that the global information and local properties captured by all layers could be exploited.

\section{Experiments}
We evaluated the proposed SesameBERT model by conducting multiple classification tasks. For comparison with the results of a previous study on BERT \citep{Devlin2019}, we reimplemented the BERT model in TensorFlow in our experiments. \footnote{Our code will be released upon acceptance.} In addition, we set most of the parameters to be identical to those in the original BERT model, namely, batch size: 16, 32, learning rate: 5e-5, 3e-5, 2e-5, and number of epochs: 3, 4. All of the results in this paper can be replicated in no more than 12 hours by a graphics processing unit with nine GLUE datasets. We trained all of the models in the same computation environment with an NVIDIA Tesla V100 graphics processing unit.

\subsection{GLUE Datasets}
GLUE benchmark is a collection of nine natural language-understanding tasks, including question-answering, sentiment analysis, identification of textual similarities, and recognition of textual entailment \citep{Wang2019}. GLUE datasets were employed because they are sets of tools used to evaluate the performance of models for a diverse set of existing NLU tasks. The datasets and metrics used for the experiments in this study are detailed in the appendix \ref{appendix:A}.

\subsection{HANS Dataset}

We used a new evaluation set, namely the HANS dataset, to diagnose fallible structural heuristics presented in a previous study \citep{McCoy2019} based on syntactic properties. More specifically, models might apply accurate labels not based on reasoning regarding the meanings of words but rather by assuming that the $premise$ entails any $hypothesis$ whose words all appear in the premise \citep{Dasgupta2018,Naik2018}. Furthermore, an instance that contradicts the lexical overlap heuristics in MNLI is likely too rare to prevent a model from learning heuristics. Models may learn to assume that a label is contradictory whenever a negation word is contained in the premise but not the hypothesis \citep{McCoy2018}. Therefore, whether a model scored well on a given test set because it relied on heuristics can be observed. For example, BERT performed well on MNLI tasks but poorly on the HANS dataset; this finding suggested that the BERT model employs the aforementioned heuristics.\par
The main difference between the MNLI and HANS datasets is their numbers of labels. The MNLI dataset has three labels, namely Entailment, Neutral, and Contradiction. In the HANS dataset, instances labeled as Contradiction or Neutral are translated into non-entailment. Therefore, this dataset has only two labels: Entailment and Non-entailment. The HANS dataset targets three heuristics, namely Lexical overlap, Subsequence, and Constituent, with more details in appendix \ref{appendix:B}. This dataset not only serves as a tool for measuring progress in this field but also enables the visualization of interpretable shortcomings in models trained using MNLI.

\subsection{Results}
\subsubsection{GLUE Datasets Results}

\begin{table*}[htb!]
\caption{Test results in relation to the GLUE benchmark. The metrics for these tasks, shown in appendix \ref{appendix:A}, were calculated using a GLUE score. We compared our SesameBERT model with the original BERT-Base model, ELMo \citep{Peters2018} and OpenAI GPT \citep{Radford2018}. All results were obtained from \url{https://gluebenchmark.com/leaderboard}.}
\begin{center}  
\begin{tabular}{l|cccc}
  \hline\hline  
    & \bf{BiLSTM+ELMo+Attn} & \bf{OpenAI GPT} & \bf{BERT-Base} & \bf{SesameBERT} \\
  \hline
CoLA      & 33.6                      & 45.4                & 52.1               & \bf{52.7}        \\
SST-2     & 90.4                      & 91.3                & 93.5               & \bf{94.2}        \\
MRPC      & 84.4                      & 82.3                & \bf{88.9}               & \bf{88.9}                 \\
STS-B     & 72.3                      & 80.0                & \bf{85.8}               & 85.5                 \\
QQP       & 63.1                      & 70.3                & \bf{71.2}               & 70.8                 \\
MNLI-m & 74.1               & 82.1           & \textbf{84.6}          & 83.7   \\  
MNLI-mm & 74.5                 &81.4           &  83.4          & \bf{83.6}   \\  
QNLI      & 79.8                      & 88.1                & 90.5               & \bf{91.0}        \\
RTE       & 58.9                      & 56.0                & 66.4               & \bf{67.6}        \\
AX        & 21.7                      & -                   & 34.2               & \bf{35.8}        \\
    \hline
GLUE score    & 70.0                      & 76.9                & 78.3               & \bf{78.6} \\
     \hline\hline  
 \end{tabular}  
    \label{table1}
\end{center}  
\end{table*} 

This subsection provides the experiment results of the baseline model and the models trained using our proposed method. We performed Gaussian blurring on attention outputs in the experiment. In addition, we employed a batch size of 32, learning rates of 3e-5, and 3 epochs over the data for all GLUE tasks. We fine-tuned the SesameBERT model through 9 downstream tasks in the datasets. For each task, we performed fine-tuning alongside Gaussian blur kernel sigmas 1e-2, 1e-1, 3e-1, and 5e-1 and selected that with the most favorable performance in the dev set. Because GLUE datasets do not distribute labels for test sets, we uploaded our predictions to the GLUE server for evaluation. The results are presented in Table \ref{table1}; GLUE benchmark is provided for reference. In most tasks, our proposed method outperformed the original BERT-Base model \citep{Devlin2019}. For example, in the RTE and AX datasets, SesameBERT yielded improvements of $1.2\%$ and $1.6\%$, respectively. We conducted experiments on GLUE datasets to test the effects of Gaussian blurring alongside BERT on the value layer and context layer. Table \ref{table2} shows the degrees of accuracy in the dev set. The performance of Gaussian blurring with respect to self-attention layers varied among cases.

\begin{table}[htb!]
\caption{Performance of Gaussian blurring alongside the BERT model. The results were tested on four GLUE datasets, with accuracy as the metric.}
\begin{center}
\begin{tabular}{l|cccc}
  \hline\hline
  & \bf{MRPC} & \bf{RTE}  & \bf{QNLI} & \bf{SST-2} \\
  \hline
     BERT &86.7 &65.3 &88.4 &\bf{92.7} \\
     Blur on Value layer  &\bf{86.8} &69.7 &\bf{90.9} &91.3 \\
     Blur on Context layer &86.5 &\bf{70.4} &90.8 &92.0  \\
     \hline\hline
 \end{tabular}  
    \label{table2}
\end{center}    
\end{table} 

\citet{Gong2019} demonstrated that different layers vary in terms of their abilities to distinguish and capture neighboring positions and global dependency between words. We evaluated the weights learned from all layers. These weights indicated that a heavier weight represents greater importance. The results are shown in appendix \ref{appendix:C}. Because the lower layer represents word embeddings that are deficient in terms of context \citep{Yang2018}, the self-attention model in the lower layer may need to encode representations with global context and may struggle to learn localness. Table \ref{table3} shows the degree of accuracy predicted by each extracted attention output layer method. The results indicated that the lower layers had lower accuracy. \par

We performed three ablation studies. First, we examined the performance of our method without blurring; we observed that Squeeze and Excitation helped the higher layer. This trend suggested that higher layers benefit more than do lower layers from Squeeze and Excitation. Second, we analyzed the effect of Gaussian blurring on the context layer. The results revealed that the method with blurring achieved higher accuracy in lower layers. We assumed that capturing short-range dependencies among neighboring words in lower layers is an effective strategy. Even if self-attention models capture long-range dependencies beyond phrase boundaries in higher layers, modeling localness remains a helpful metric. Finally, we observed the direct effects of SesameBERT. Although our proposed architecture performed poorly in lower layers, it outperformed the other methods in higher layers. This finding indicated that in higher layers, using Squeeze and Excitation alongside Gaussian blurring helps self-attention models to capture global information in all layers.

\begin{table*}[htb!]
\caption{Comparison of specified layers among various approaches in the RTE dataset. We dissected our models into two methods. SE-BERT refers to BERT with Squeeze and Excitation; BLUR-BERT refers to BERT with Gaussian blurring.}
\begin{center}
\begin{tabular}{l| cccc}
  \hline\hline
  \bf {Layers}  &\bf {BERT}  &\bf {SE-BERT}
  &\bf {BLUR-BERT} &\bf {SesameBERT} \\
  \hline
  \multicolumn{5}{c}{Dev Set Accuracy} \\
  \hline
     First Hidden Layer   &58.1 &57.0 &\bf{64.6} &54.5 \\
     Second Hidden Layer   &55.6 & 56.3 &\bf{57.4} &54.2 \\
     \hline
     Second-to-Last Hidden  &64.6 &69.0 &67.5 &\bf{70.8} \\  
     Last Hidden    &65.3 &67.9 &68.6 &\bf{70.4}      \\ 
     \hline 
     Sum Last Four Hidden  &65.0 &69.3 &68.2  &\bf{69.7}    \\   
     Sum All 12 Layers     &68.2 & 68.6 &66.4  &\bf{69.0}  \\
     Weighted Average Layers &66.8 & 67.5 &67.9  &\bf{70.0}  \\
     \hline\hline 
 \end{tabular}  
    \label{table3}
\end{center}
\end{table*}

\subsubsection{HANS Dataset Results}
\begin{figure}[htb!]
        \begin{center}
        \includegraphics[width=\linewidth]{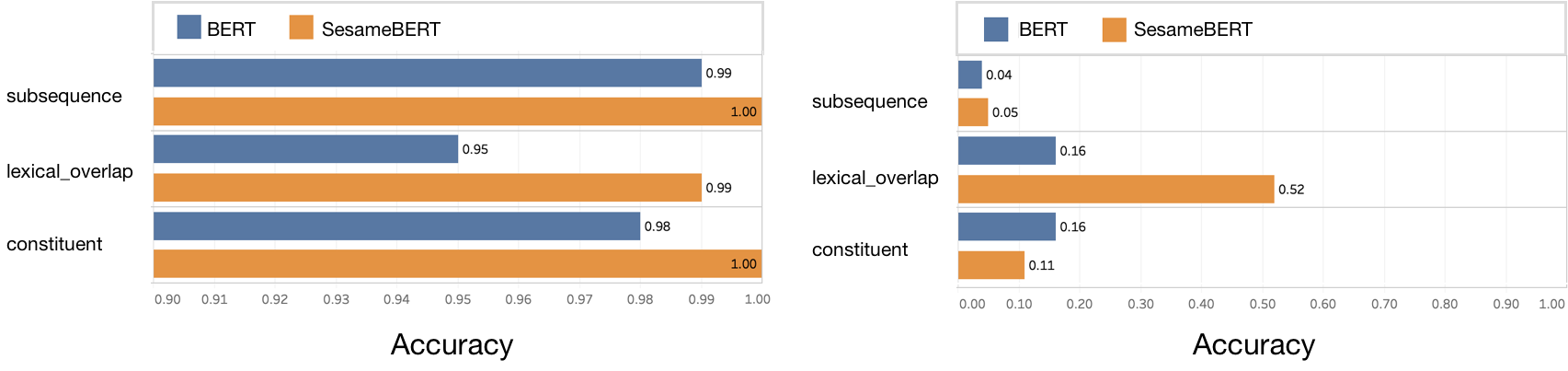}
        \caption{\label{Figure3}We compared BERT and SesameBERT for each case. \textbf{Left}: Results of heuristics-entailed cases. \textbf{Right}: Results of heuristics labeled as Nonentailment. In contrast to the results in \textbf{Left}:, BERT performed poorly in all three cases in \textbf{Right}; this indicated that the model had adopted shallow heuristics rather than learning the latent information that it intended to capture.}
        \end{center}
\end{figure}

We trained both BERT and SesameBERT on the MNLI-m dataset to evaluate their classification accuracy. Similar to the results of another study \citep{Devlin2019}, BERT achieved $84.6\%$ accuracy, which is higher than that of SesameBERT, as shown in Table \ref{table1}. In the HANS dataset, we explored the effects of two models on each type of heuristic. The results are presented in Figure \ref{Figure3}; we first examined heuristics for which the label was Entailment. We can see that both models performed well; they assigned the correct labels almost $100\%$ of the time, as we had expected them to do after adopting the heuristics targeted by HANS.
\par
Next, we evaluated the heuristics labeled as Non-entailment. BERT performed poorly for all three cases, meaning that BERT assigned correct labels based on heuristics instead of applying the correct rules of inference. By contrast, our proposed method performed almost three times as well as BERT in the case of "Lexical overlap". 

This paper argues that capturing local contexts for self-attention networks with Gaussian blurring can prevent models from easily adopting heuristics. Although our models performed poorly in cases of "Subsequence" and "Constituent", both of these heuristics may be hierarchical cases of the lexical overlap heuristic, meaning that the performance of this hierarchy would not necessarily match the performance of our models \citep{McCoy2019}.

\section{Conclusion}
This paper proposes a fine-tuning approach named SesameBERT based on the pretraining model BERT to improve the performance of self-attention networks. Specifically, we aimed to find high-quality attention output layers and then extract information from aspects in all layers through Squeeze and Excitation. Additionally, we adopted Gaussian blurring to help capture local contexts. Experiments using GLUE datasets revealed that SesameBERT outperformed the BERT baseline model. The results also revealed the weight distributions of different layers and the effects of applying different Gaussian-blurring approaches when training the model. Finally, we used the HANS dataset to determine whether our models were learning what we wanted them to learn rather than using shallow heuristics. We highlighted the use of lexical overlap heuristics as an advantage over the BERT model. SesameBERT could be further applied to prevent models from easily adopting shallow heuristics.

\bibliography{iclr2020_conference}

\begin{thebibliography}{20}
\providecommand{\natexlab}[1]{#1}
\providecommand{\url}[1]{\texttt{#1}}
\expandafter\ifx\csname urlstyle\endcsname\relax
  \providecommand{\doi}[1]{doi: #1}\else
  \providecommand{\doi}{doi: \begingroup \urlstyle{rm}\Url}\fi

\bibitem[Bahdanau et~al.(2015)Bahdanau, Cho, and Bengio]{Bahdanau2015}
Dzmitry Bahdanau, Kyunghyun Cho, and Yoshua Bengio.
\newblock Neural machine translation by jointly learning to align and
  translate.
\newblock \emph{ICLR}, 2015.

\bibitem[Baosong~Yang(2018)]{Yang2018}
Derek F. Wong Fandong Meng Lidia S. Chao Tong~Zhang Baosong~Yang, Zhaopeng~Tu.
\newblock Modeling localness for self-attention networks.
\newblock \emph{EMNLP}, 2018.

\bibitem[Carmona et~al.(2018)Carmona, Mitchell, and Riedel]{Carmona2018}
V.~Ivan~Sanchez Carmona, Jeff Mitchell, and Sebastian Riedel.
\newblock Behavior analysis of nli models: Uncovering the influence of three
  factors on robustness.
\newblock \emph{NAACL}, 2018.

\bibitem[Dasgupta et~al.(2018)Dasgupta, Guo, Stuhlmuller, Gershman, and
  Goodman]{Dasgupta2018}
Ishita Dasgupta, Demi Guo, Andreas Stuhlmuller, Samuel~J. Gershman, and Noah~D.
  Goodman.
\newblock Evaluating compositionality in sentence embeddings.
\newblock 2018.

\bibitem[Devlin et~al.(2019)Devlin, Chang, Lee, and Toutanova]{Devlin2019}
Jacob Devlin, Ming-Wei Chang, Kenton Lee, and Kristina Toutanova.
\newblock Bert: Pre-training of deep bidirectional transformers for language
  understanding.
\newblock \emph{NAACL}, 2019.

\bibitem[Gong et~al.(2019)Gong, He, Li, Qin, Wang, and Liu]{Gong2019}
Linyuan Gong, Di~He, Zhuohan Li, Tao Qin, Liwei Wang, and Tieyan Liu.
\newblock Efficient training of bert by progressively stacking.
\newblock \emph{ICML}, 2019.

\bibitem[Hu et~al.(2018)Hu, Shen, Albanie, Sun, and Wu]{Hu2017}
Jie Hu, Li~Shen, Samuel Albanie, Gang Sun, and Enhua Wu.
\newblock Squeeze-and-excitation networks.
\newblock \emph{CVPR}, 2018.

\bibitem[Li et~al.(2019)Li, Yang, Dou, Wang, Lyu, and Tu]{Li2019}
Jian Li, Baosong Yang, Zi-Yi Dou, Xing Wang, Michael~R. Lyu, and Zhaopeng Tu.
\newblock Information aggregation for multi-head attention with
  routing-by-agreement.
\newblock \emph{NAACL}, 2019.

\bibitem[McCoy \& Linzen(2019)McCoy and Linzen]{McCoy2018}
R.~Thomas McCoy and Tal Linzen.
\newblock Non-entailed subsequences as a challenge for natural language
  inference.
\newblock \emph{SCiL}, 2019.

\bibitem[McCoy et~al.(2019)McCoy, Pavlick, and Linzen]{McCoy2019}
R.~Thomas McCoy, Ellie Pavlick, and Tal Linzen.
\newblock Right for the wrong reasons: Diagnosing syntactic heuristics in
  natural language inference.
\newblock \emph{ACL}, 2019.

\bibitem[Mikolov et~al.(2013)Mikolov, Sutskever, Chen, Corrado, and
  Dean]{Mikolov2013}
Tomas Mikolov, Ilya Sutskever, Kai Chen, Greg Corrado, and Jeffrey Dean.
\newblock Distributed representations of words and phrases and their
  compositionality.
\newblock \emph{NIPS}, 2013.

\bibitem[Miller et~al.(2016)Miller, Fisch, Dodge, Karimi, Bordes, and
  Weston]{Miller2016}
Alexander~H. Miller, Adam Fisch, Jesse Dodge, Amir-Hossein Karimi, Antoine
  Bordes, and Jason Weston.
\newblock Key-value memory networks for directly reading documents.
\newblock 2016.

\bibitem[Naik et~al.(2018)Naik, Ravichander, Sadeh, Rose, and Neubig]{Naik2018}
Aakanksha Naik, Abhilasha Ravichander, Norman Sadeh, Carolyn Rose, and Graham
  Neubig.
\newblock Stress test evaluation for natural language inference.
\newblock \emph{COLING}, 2018.

\bibitem[Pennington et~al.(2014)Pennington, Socher, and
  Manning]{Pennington2014}
Jeffrey Pennington, Richard Socher, and Christopher~D. Manning.
\newblock Glove: Global vectors for word representation.
\newblock \emph{EMNLP}, 2014.

\bibitem[Peters et~al.(2018)Peters, Neumann, Iyyer, Gardner, Clark, Lee, and
  Zettlemoyer]{Peters2018}
Matthew~E. Peters, Mark Neumann, Mohit Iyyer, Matt Gardner, Christopher Clark,
  Kenton Lee, and Luke Zettlemoyer.
\newblock Deep contextualized word representations.
\newblock \emph{NAACL}, 2018.

\bibitem[Radford et~al.(2018)Radford, Narasimhan, Salimans, and
  Sutskever]{Radford2018}
Alec Radford, Karthik Narasimhan, Tim Salimans, and Ilya Sutskever.
\newblock Improving language understanding by generative pre-training.
\newblock 2018.

\bibitem[Vaswani et~al.(2017)Vaswani, Shazeer, Parmar, Uszkoreit, Jones, Gomez,
  Kaiser, and Polosukhin]{Vaswani2017}
Ashish Vaswani, Noam Shazeer, Niki Parmar, Jakob Uszkoreit, Llion Jones,
  Aidan~N. Gomez, Lukasz Kaiser, and Illia Polosukhin.
\newblock Attention is all you need.
\newblock \emph{NIPS}, 2017.

\bibitem[Wang et~al.(2019)Wang, Singh, Michael, Hill, Levy, and
  Bowman]{Wang2019}
Alex Wang, Amanpreet Singh, Julian Michael, Felix Hill, Omer Levy, and
  Samuel~R. Bowman.
\newblock Glue: A multi-task benchmark and analysis platform for natural
  language understanding.
\newblock \emph{ICLR}, 2019.

\bibitem[Wang et~al.(2017)Wang, Zhang, Xie, Zhou, Premachandran, Zhu, Xie, and
  Yuille]{Wang2017}
Jianyu Wang, Zhishuai Zhang, Cihang Xie, Yuyin Zhou, Vittal Premachandran, Jun
  Zhu, Lingxi Xie, and Alan Yuille.
\newblock Visual concepts and compositional voting.
\newblock \emph{Annals of Mathematical Sciences and Applications}, 2017.

\bibitem[Yang et~al.(2019)Yang, Wang, Wong, Chao, and Tu]{Yang2019}
Baosong Yang, Longyue Wang, Derek Wong, Lidia~S. Chao, and Zhaopeng Tu.
\newblock Convolutional self-attention networks.
\newblock \emph{NAACL}, 2019.

\end{thebibliography}
\bibliographystyle{iclr2020_conference}

\newpage
\appendix
\section{Descriptions of GLUE DATASETS}
\label{appendix:A}

\begin{table*}[htb!]
\caption{Descriptions of GLUE tasks. The second and third column denote the sizes of the corresponding corpora. All tasks are classification tasks, except for STS-B, which is a regression task.}
\begin{small} 
\begin{center}
\begin{tabular}{l|cccc|c}  
  \hline\hline
   \bf{Corpus} & \bf{\#Train}   & \bf{\#Test}
  & \bf{Task}  & \bf{Metrics}  & \bf{Domain}\\
  \hline
  \multicolumn{6}{c}{$Single-Sentence\  Tasks$} \\
  \hline
     CoLA   &8.5k &1k &acceptability   &Matthews correlation  &misc \\
     SST-2  &67k &1.8k &sentiment  &Accuracy  &movie reviews \\  
     \hline 
     \multicolumn{6}{c}{$Similarity/Paraphrase\ Tasks$} \\
     \hline
     MRPC    &3.7k &1.7k &paraphrase  &Accuracy/F1  &news     \\   
     STS-B   &7k &1.4k &sentence similarity  &Pearson/Spearman corr.   &misc     \\   
     QQP     &364k &391k&paraphrase  &Accuracy/F1  &social QA  \\
     \hline
     \multicolumn{6}{c}{$Inference\ Tasks$} \\
     \hline
     MNLI    &393k &20k&NLI  & Accuracy  &misc     \\   
     QNLI    &105k &5.5k &QA/NLI  &Accuracy &Wikipedia     \\   
     RTE    &2.5k &3k &NLI  &Accuracy &Wikipedia     \\ 
     AX     &- &1.1k & NLI &Matthews correlation  & news, paper, etc   \\ 
     \hline\hline 

 \end{tabular}  
    \end{center}
    \end{small}  
\end{table*}

\section{Description of HANS DATASET}
\label{appendix:B}

\begin{table*}[htb!]
\caption{Three types of heuristics targeted by the HANS dataset. The examples show incorrect entailment predictions  that would result from targeting these heuristics.}
\begin{small} 
\begin{center}
\begin{tabular}{p{2cm} p{6cm} p{4.5cm} }  
  \hline\hline 
   Heuristic   & Definition  &  Example \\
  \hline
     Lexical overlap    &  Assume that a premise entails all hypotheses constructed from words in the premise   & \textbf{The docter} was \textbf{paid} by \textbf{the actor}. $\xrightarrow[\text{WRONG}]{}$ The doctor paid the actor.  \\
     \hline 
     Subsequence    &Assume that a premise entails all of its
contiguous subsequences.  &The doctor near \textbf{the actor danced}. $\xrightarrow[\text{WRONG}]{}$ The actor danced.       \\   
     \hline
     Constituent    &Assume that a premise entails all complete
subtrees in its parse tree.   &If \textbf{the artist} slept, the actor ran. $\xrightarrow[\text{WRONG}]{}$ The artist slept.     \\  
     \hline\hline 
 \end{tabular}  
    \end{center}
    \end{small}  
\end{table*} 

\section{Layer weights calculated by Squeeze and Excitation}
\label{appendix:C}
\begin{figure*}[b]
        \begin{center}
        \includegraphics[scale=0.25]{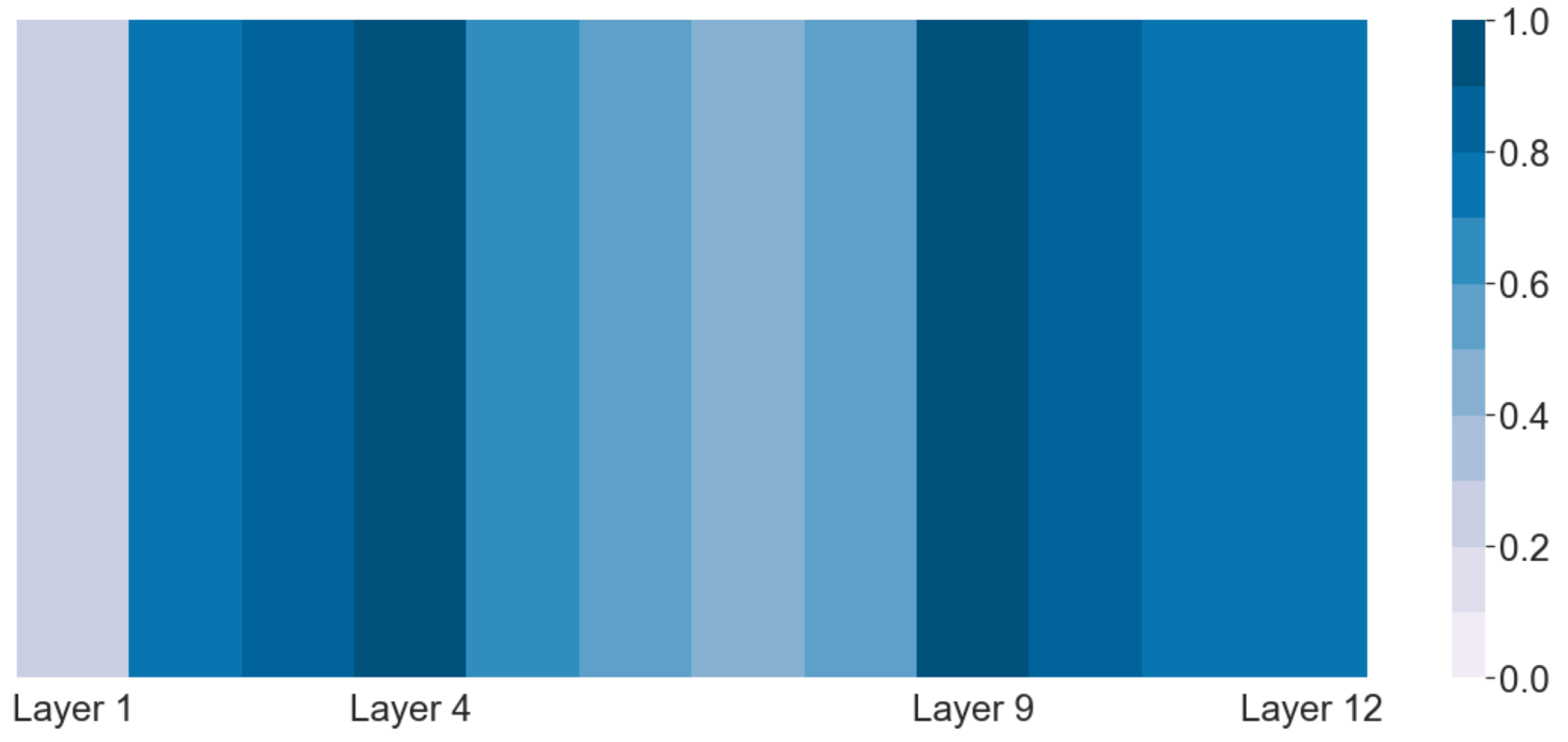}
        \caption{Evaluation of the weights calculated by Squeeze and Excitation for all layers, with the RTE dataset as an example.}
        \end{center}
\end{figure*}

\end{document}